\documentclass[9pt]{article}
\usepackage{amsmath,graphicx,mlspconf}
\usepackage{booktabs}
\usepackage{multirow}
\usepackage{subcaption}
\usepackage{array,makecell}
\usepackage{comment}
\usepackage{placeins}
\usepackage{algorithm}
\usepackage{algorithmic}

\toappear{Preprint to appear at IEEE MLSP 2024}

\copyrightnotice{            \begin{tabular}{l}
{\copyright} 20XX IEEE. Personal use of this material is permitted. Permission from IEEE must be obtained for all other uses, in any current or future media,\\
including reprinting/republishing this material for advertising or promotional purposes, creating new collective works, for resale or redistribution to\\
servers or lists, or reuse of any copyrighted component of this work in other works.
            \end{tabular}}

\title{Visualize and Paint GAN Activations}
\name{Rudolf Herdt, Peter Maass}
\address{Center for Industrial Mathematics, University of Bremen, Bremen 28334, Germany}

\begin{document}

\maketitle

\begin{abstract}
We investigate how generated structures of GANs correlate with their activations in hidden layers, with the purpose of better understanding the inner workings of those models and being able to paint structures with unconditionally trained GANs. This gives us more control over the generated images, allowing to generate them from a semantic segmentation map while not requiring such a segmentation in the training data. To this end we introduce the concept of tileable features, allowing us to identify activations that work well for painting.\footnote{The code is available at: https://github.com/rherdt185/visualize-gan-activations}
%
%
%
%
%
%
\end{abstract}

%

\section{Introduction}

This paper is motivated by our research on digital pathology and in particular by the task to develop reliable tools for the segmentation of stained whole slide
tissue images into healthy and tumorous regions. 
In this context, the availability of sufficient training data is crucial and GANs can be used to generate additional data for different tumor entities.
 
For a better understanding of this concept we aim to visualize activation vectors of
these GANs, similar to what was already done for classification
models via gradient descent [1]. 
%
However, there are certain differences, when comparing the classical classification task with our image generation problem. 
Visualization
of activations for classification models typically aims to
find an input that maximizes a particular channel in a hidden
layer, e.g. a prototype for a particular class. However, 
when using GANs for data generation it is the output of the model that is interpretable and human
understandable, not the input. 
This gives us the advantage, that we can visualize activations by a single forward
pass, instead of using iterative gradient descent to synthesize
an input. This allows generating visualization interactively, in
real-time. 
Furthermore, we focus on visualizing individual vectors
(in the following we refer to them as activation vectors), and
not channels. In the classification setting, that would correspond
to finding an input where the mean activations (over
height and width) are as close as possible to the activation
vector.
Once we have determined suitable activation vectors in hidden layers, which correspond to certain output structures, we can use them in two ways for visualization.
As a first try,
we set all spatial activations to the value of the activation vector
we want to visualize. While this works well for some activation
vectors (we call them tileable), it fails for others which
need contrast in the activations (we call those non-tileable).
As a second approach we can embed these activation vectors at certain spatial locations in the hidden layer and generate artificial images with naturally looking but spatially restricted structures.

In this paper
we investigate and measure different methods for visualizing
activation vectors in 3 StyleGAN2 models trained on AFHQ
Wild, BreCaHAD and LSUN Church data, and for a custom
GAN trained on in-house real-world digital pathology scans
of stained tissue samples.
Further, generating annotated data for semantic segmentation
is a time consuming process. Therefore, we introduce the concept of tileable activations and investigate
painting with activation vectors in hidden layers of the GAN,
which could be used to generate annotated data for semantic
segmentation.

\begin{figure*}[tb]
\begin{center}
\centerline{\includegraphics[width=0.8\textwidth]{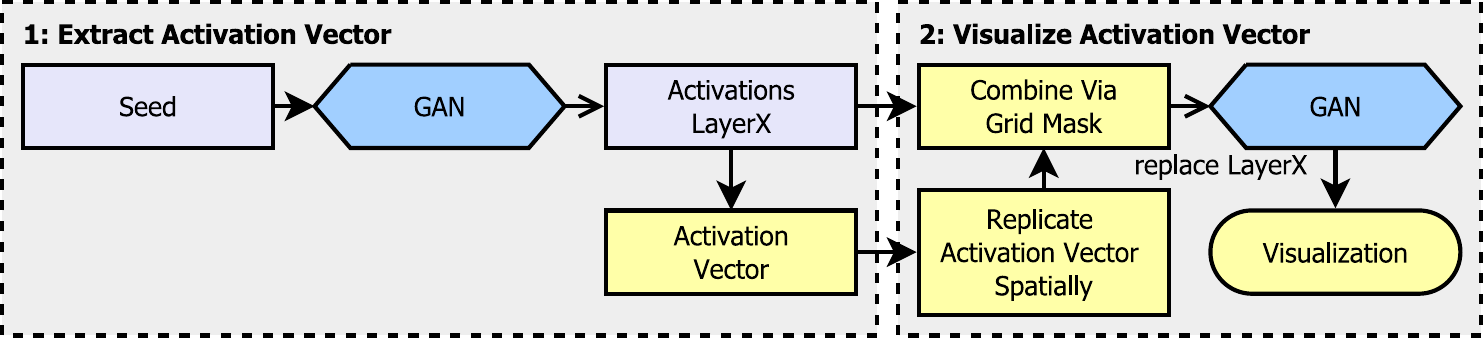}}
\caption{Overview of our visualization method}
\label{fig:visualization_methods}
\end{center}
\vskip -0.2in
\end{figure*}

\section{Related Work}

\subsection{Visualize Activation Vectors in Classification Models}

This work is related to existing work focusing on visualizing activation vectors of classification models, which is done by inverting them via gradient descent \cite{olah2017feature}.
%
%
%
That approach can be used to visualize specific activation vectors \cite{carter2019activation}, and is not limited to convolutional neural networks, but can also be used for vision transformer \cite{ghiasi2022vision}.

In our case, we visualize activation vectors for a GAN model, which we do via a single forward pass, since here we have the visualization in the output layer of the GAN (as opposed to the input layer for the classification models).
This allows us to generate visualizations interactively in real-time.
Further, recent work has shown that pretrained classification models and GANs can be stitched in a hidden layer into an autoencoder via a single linear transformation (a trained 1x1 convolution) \cite{pmlr-v222-herdt24a}.
This suggests that the insights that can be gained from interpreting GANs, can transfer over to classification models.

\subsection{Investigate GAN Channels and Painting}

%
Earlier efforts in understanding GAN models focused on analysing the meaning of channels in hidden layers, by segmenting the generated images with a segmentation model \cite{GANDissect}.
Then they compared the activations of the GAN channels with the segmentation masks predicted from the segmentation model for the different classes (e.g. which GAN channels activate where the segmentation model predicts trees, or in other words, which channels are responsible for generating trees).
They double checked this by ablating or increasing the activations of those channels and observe the change in the generated image.

They also painted with structures in the generated image, by painting with those channels in hidden layer.
But the painting did not always succeed, it depended on context (e.g. they could not paint a door into the sky).
In contrast, in our case we visualize and paint with whole activation vectors (using all channels), instead of sets of channels.
This allows us to paint the structures into any context.
For the digipath GAN the usage of painting is similar to GANs conditioned on semantic segmentation masks (like \cite{park2019SPADE}), but does not require such annotated training data.
Instead we identify activation vectors that represent the desired structures, and by painting with them in hidden layers we paint with the structure in the generated image.

\subsection{Analyse Vectors in Latent Space}

GANs typically work by generating a sample starting from a single vector.
This vector is eventually reshaped to have a spatial size and then subsequently upscaled to the final output resolution.
While we only focus on the hidden layers that already have a spatial size, there is previous work exploring the earlier latent space where an image is represented by a single vector.
For example Radford \cite{DBLP:journals/corr/RadfordMC15} explored this latent space by doing vector arithmetic and interpolation between such vectors.
%
%
For a GAN that generates synthetic images of faces, they observed e.g. that taking a vector where the GAN generates an image of a man with glasses, subtracting a vector for a man without glasses and adding a vector for a women results in an image of a women with glasses.

\subsection{Visualize GAN Training}

Other work focuses on explaining the behavior of GANs by interactively training a simple GAN \cite{Kahng2018GANLU}.
All spaces in the investigated GAN are two dimensional, which allows to directly plot them without dimensionality reduction.
This allows to visualize how the GAN maps from its input to output, how the hidden layers transform the latent spaces.

\section{Methods}
\label{sec:methods}

\begin{figure}[tb]
    \centering
    \begin{subfigure}{0.15\textwidth}
        \centering
        \includegraphics[width=\textwidth]{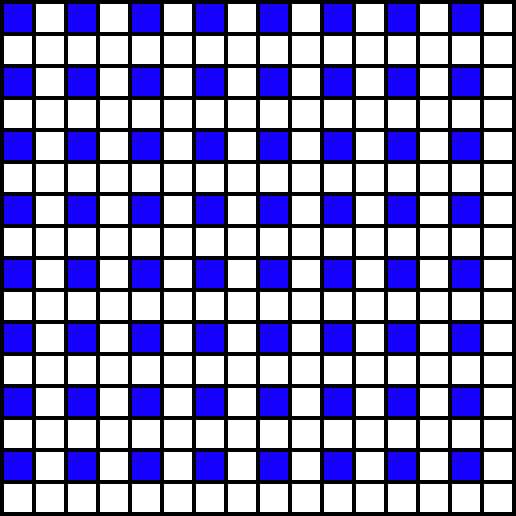}
        \label{fig:gsub1}
    \end{subfigure}
    \begin{subfigure}{0.15\textwidth}
        \centering
        \includegraphics[width=\textwidth]{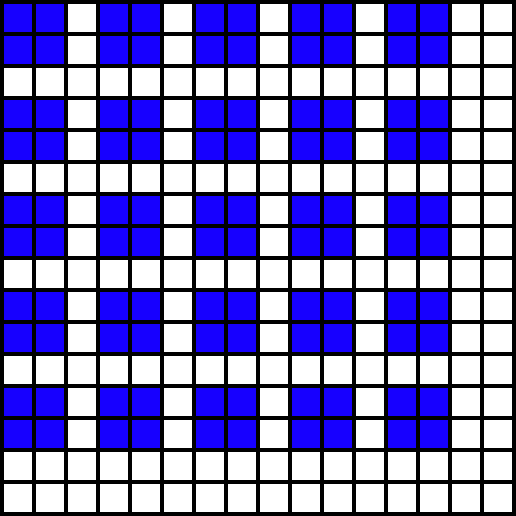}
        \label{fig:gsub2}
    \end{subfigure}
    \begin{subfigure}{0.15\textwidth}
        \centering
        \includegraphics[width=\textwidth]{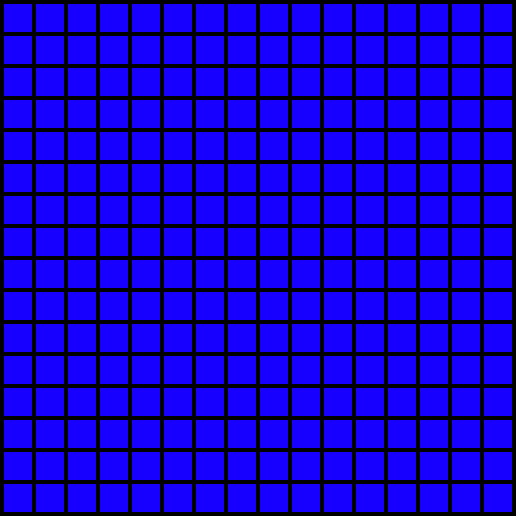}
        \label{fig:gsub3}
    \end{subfigure}

    \vskip -0.2in
    \caption{Example grid masks. From left to right is: Using a grid size of 1; Using a grid size of 2; Fully replicating the activation vector}
    \label{fig:grid_size_visualization}
\end{figure}

In this section we describe our methods for visualizing and painting with activation vectors.

\subsection{Visualization}

An overview of our visualization method is shown in Figure \ref{fig:visualization_methods}.
It consists of two subsequent steps, first getting an activation vector and second visualize that activation vector.
In the following we describe both steps in more detail.


%
%
In the first step we extract an activation vector $v$ out of a hidden layer of the GAN (in the following the layer is denoted as LayerX).
Simply sampling a random vector from e.g. Gaussian noise will likely be out of distribution and will therefore not be a 'valid' activation vector.
Instead we can get an activation vector $v$, by extracting a pixel of the activations at LayerX.


In the second step we visualize the activation vector $v$.
For a GAN we can understand and interpret the output layer.
Therefore, we want to find an output that illustrates $v$.
%
As a first try, we wanted the output to be only dependent on $v$ and not on any other activations.
%
%
Therefore we override the activations in LayerX, by setting all pixels to $v$ (replicate the activation vector $v$ spatially).
The output of the GAN is then the desired visualization.

While this works well for some activation vectors (in the following we refer to those as tileable) where the generated structure tiles, like for e.g. fur, it fails for others.
For those other activation vectors the generated structure does not tile, but stretch, eventually looking unrealistic (in the following we refer to those as non-tileable).
For those non-tileable ones we cannot set all spatial activations to $v$ (because this results in a stretched structure looking unrealistic).
%
%
%
For them we set the activations to $v$ based on a grid, having different activations in between the grid blocks.
Figure \ref{fig:grid_size_visualization} shows the grid for a grid size of 1 and 2 (left and middle image) and in the right image we would set all spatial activations to $v$.
All the activations at the blue pixels will be set to $v$, whereas the white pixels would keep their original activations.

\subsection{Painting with Activation Vectors}

\begin{figure}[tb]
\begin{center}
\centerline{\includegraphics[width=0.17\textwidth]{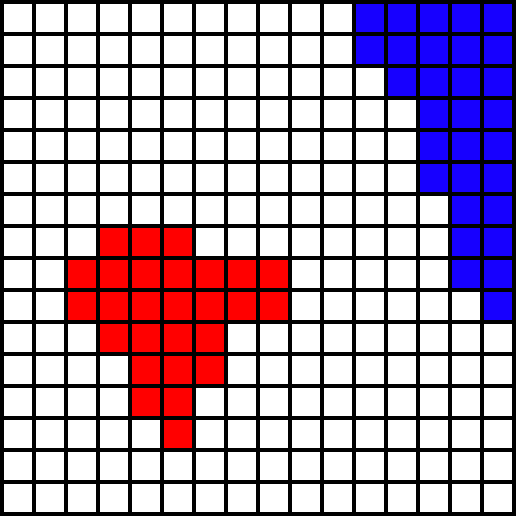}}
\caption{Example painting}
\label{fig:painting_example}
\end{center}
\vskip -0.4in
\end{figure}

In this section we describe our method for painting with activation vectors.
From visualizing activation vectors, we can see human understandable structures, like fur, eyes, or different kinds of tissue.
This brings up the question, whether painting with activation vectors in hidden layers of the GAN results in those structures in the generated image.
I.e. whether painting with an activation vector where we see eyes in the visualization allows us to paint eyes in the generated image.

For painting with activation vectors we simply replace pixels in the hidden layer we are painting at, with the activation vector we wish to paint with.
An example is illustrated in Figure \ref{fig:painting_example}, where we would paint with two activation vectors.
All blue pixels would be replaced with one activation vector, all red ones with the other.
The white pixels would be kept at the original activations, they would remain unchanged.

\section{Experiments}
\label{sec:experiments}

In this section we report the setup and results from our experiments.
We start with the datasets and models we utilized. 
%
Then we describe our experiment for separating between tileable and non-tileable features and investigate the optimal grid size when visualizing activation vectors.
Afterwards we discuss a case study for painting with activation vectors, in order to generate annotated data for semantic segmentation in tumor detection in digital pathology.
Finally we discuss the limitations of our methods.
%

\subsection{Datasets and Models}

\begin{table}[tb]
    \caption{GANs and image sizes used in the experiments}
    \label{tab:datasets_and_models}
    \begin{center}
    \begin{small}
    \begin{sc}
    \vskip -0.2in
    \resizebox{.5\textwidth}{!}{
    \begin{tabular}{l|cccc}  
    \toprule
    Dataset & GAN & \makecell{Feature \\ Extractor} & Image Size & Resized To \\
    \midrule
    AFHQ Wild   & StyleGAN2 & ResNet50 & 512x512 & 256x256 \\
    LSUN Churches & StyleGAN2 & ResNet50 & 256x256 & 256x256 \\  
    BreCaHAD & StyleGAN2 & ResNet18 & 512x512 & 256x256 \\  
    Digipath & Custom & ResNet34 & 1536x1536 & 1536x1536 \\ 
    \bottomrule
    \end{tabular}
    }
    \end{sc}
    \end{small}
    \end{center}
    \vskip -0.1in
\end{table}

\begin{figure}[tb]
\begin{center}
\centerline{\includegraphics[width=0.5\textwidth]{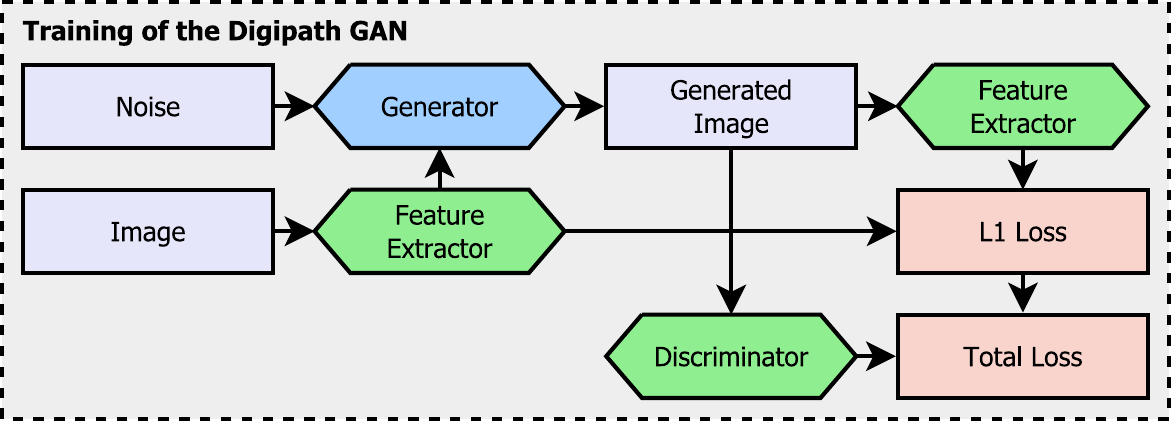}}
\caption{Using perceptual loss from a feature extractor to help guide the training of the digipath GAN}
\label{fig:digipath_gan}
\end{center}
\vskip -0.2in
\end{figure}

Table \ref{tab:datasets_and_models} shows the datasets and models we use in our experiments.
As GAN models we use 3 pretrained StyleGAN2 \cite{karras2020stylegan2} models, which are openly available.
Additionally we use a GAN (in the following referred to as digipath GAN) trained on in-house scanned histological tissue samples.
As feature extractors we use a ResNet50 \cite{resnets} available from torchvision model zoo for the AFHQ Wild and LSUN Church GANs and a ResNet18 available from monai model zoo for the BreCaHAD GAN.
%
%


In the following we describe the digipath GAN in more detail.
We use a UNet (including skip connections) as architecture for this GAN.
As input it gets Gaussian noise at a resolution of 1536x1536, which is subsequently downsampled to a size of 3x3 and then upsampled to a size of 1536x1536 again.
Additionally we use a feature extractor (a ResNet34 encoder of a segmentation model trained to detect different types of tumor) to guide the training process, as shown in Figure \ref{fig:digipath_gan}.
Each training image gets passed into the ResNet34, and we extract its activations at Layer3 and Layer4.
Then we compute the mean over height and width and concatenate the two resulting vectors into one.
This vector is replicated to a spatial size of 3x3 and concatenated with the activations of the GAN generator at the deepest point in the UNet.
If we then pass the generated image into the feature extractor again, we want to arrive at the same vector the GAN started from, and use L1 loss between the two vectors.
This additional loss helped us to stabilize training of the GAN, and we can still get image variations by varying the noise input.
But this also means that in actual usage of the GAN we also need to have such vectors, and cannot sample them from a noise distribution.
For our experiments we have generated vectors from 24522 images, and for generating an image with the GAN we sample a random vector from those 24522.
%

\begin{figure}[tb]
\begin{center}
\centerline{\includegraphics[width=0.5\textwidth]{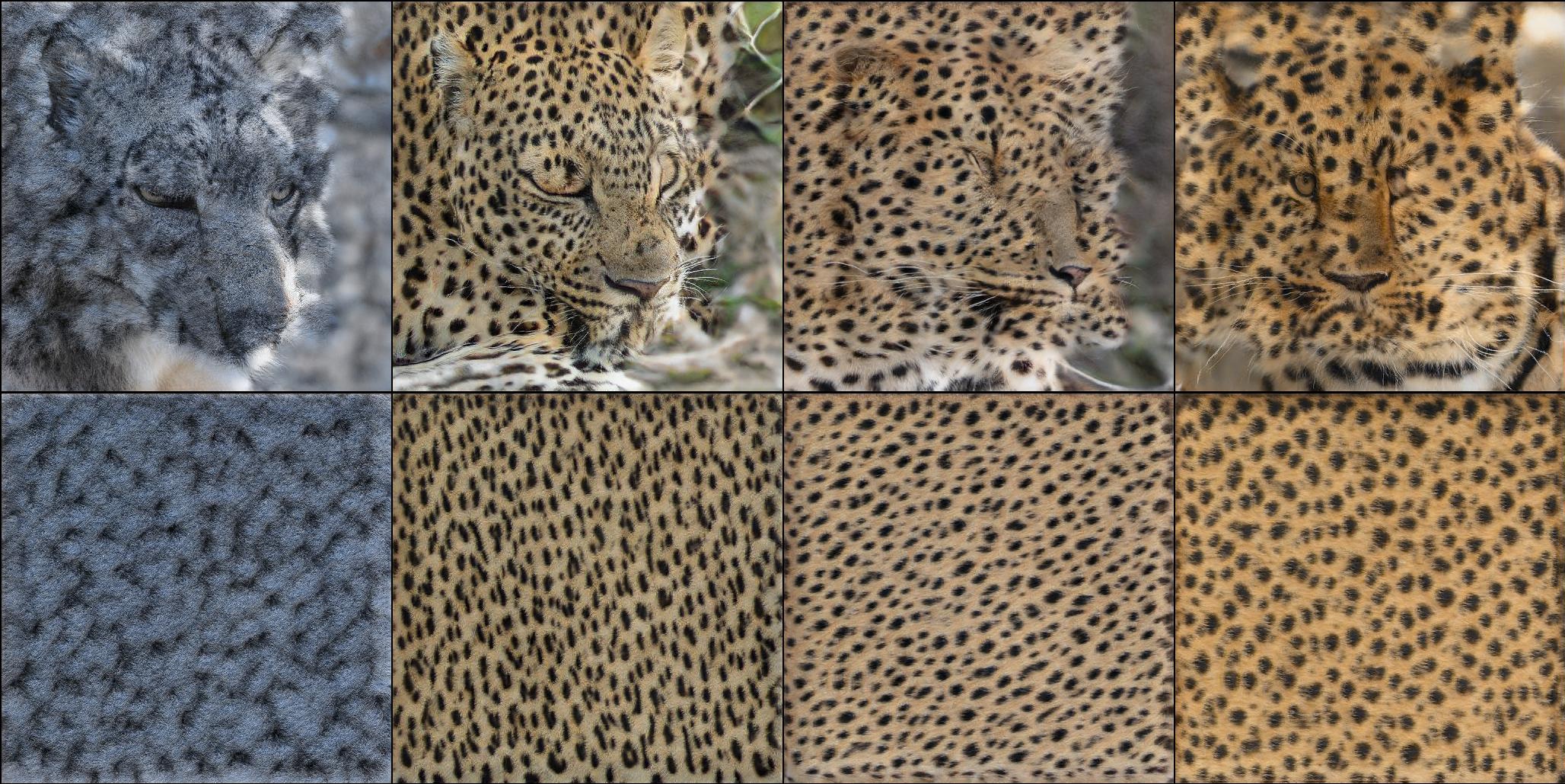}}
\caption{4 visualizations with the highest cosine similarity, i.e. tileable features}
\label{fig:highest_cossim}
\end{center}
\vskip -0.2in
\end{figure}

\subsection{Tileable and Non-Tileable Features}

In this section we want to find a way to separate between tileable and non-tileable features.

\subsubsection{Examples}

\begin{figure}[tb]
\begin{center}
\centerline{\includegraphics[width=0.5\textwidth]{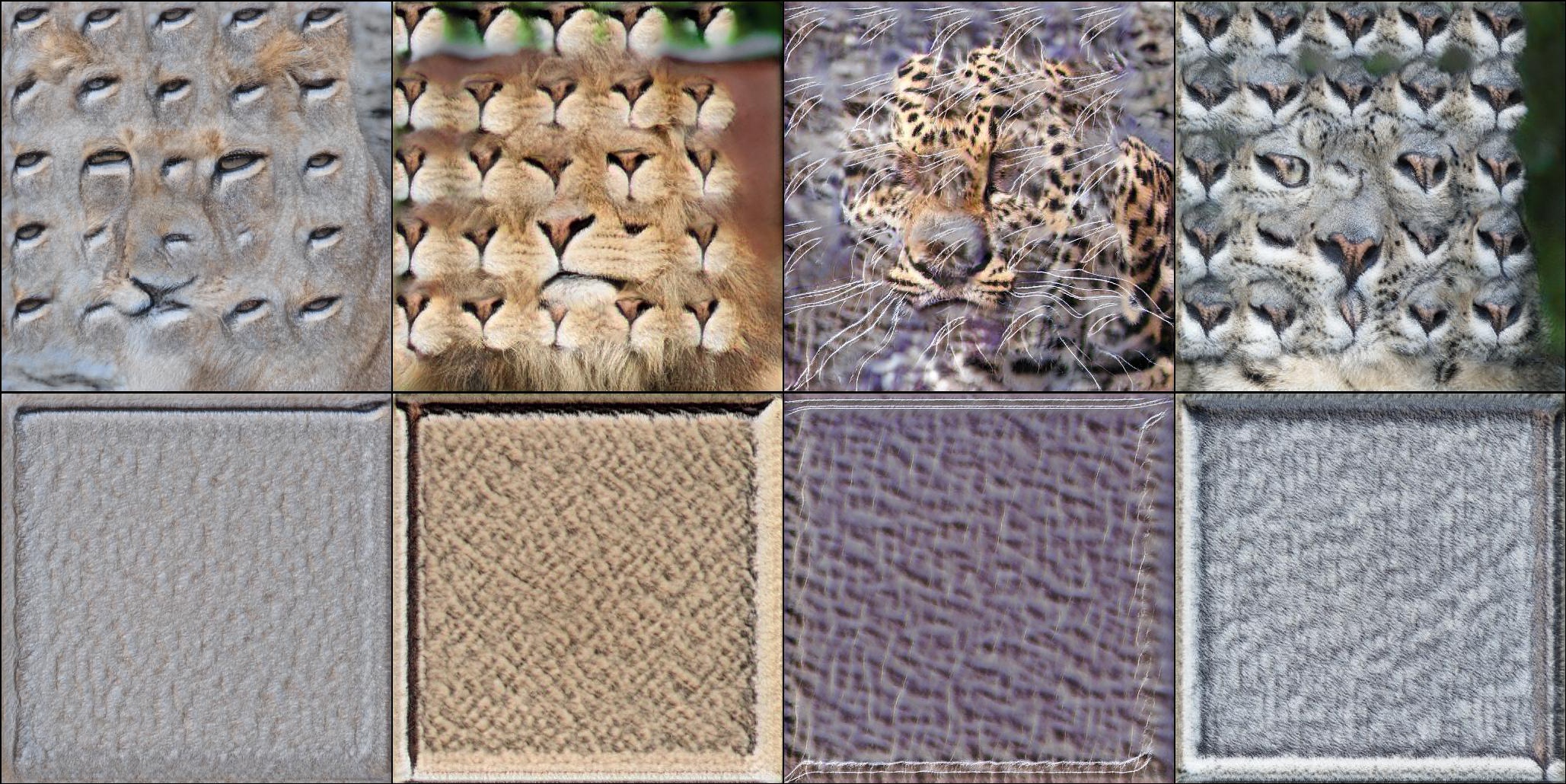}}
\caption{4 visualizations with the lowest cosine similarity, i.e. non-tileable features}
\label{fig:lowest_cossim}
\end{center}
\vskip -0.2in
\end{figure}

First we look at examples for tileable and non-tileable features, in order to illustrate a way to separate them.
Figure \ref{fig:highest_cossim} shows 4 tileable features.
In order to generate the visualizations shown in the bottom row, we set all spatial activations to $v$ (we use the mask shown in Figure \ref{fig:grid_size_visualization} on the right).
%
%
In the top row we use a grid with a size of 2 (we use the mask shown in Figure \ref{fig:grid_size_visualization} in the middle).
We can see that the top row limited to the grid mask, and the bottom row look similar.
This is contrary to the non-tileable features shown in Figure \ref{fig:lowest_cossim}.
For them the top row limited to the grid mask and the bottom row do not look similar.
For example, in the first image from the left we can see a grid of eyes in the top image.
In the bottom image however, we see only one long stretched eye at the top of the image and fur in the rest of it.

This means that for tileable features, generating visualizations while setting all spatial activations to $v$ or only in a grid results in similar structures in the generated image.
Whereas for non-tileable features we get different structures in the generated image.
This suggests that by measuring the similarity of both visualizations, we should be able to separate between tileable (those will have high similarity) and non-tileable features (those will have lower similarity).
To measure the similarity, we utilize a pretrained classification model (ResNet50 trained on ImageNet).
In the following we describe how we generate pairs of visualizations and measure their similarity.

\begin{figure*}[htb]
\begin{center}
\centerline{\includegraphics[width=0.8\textwidth]{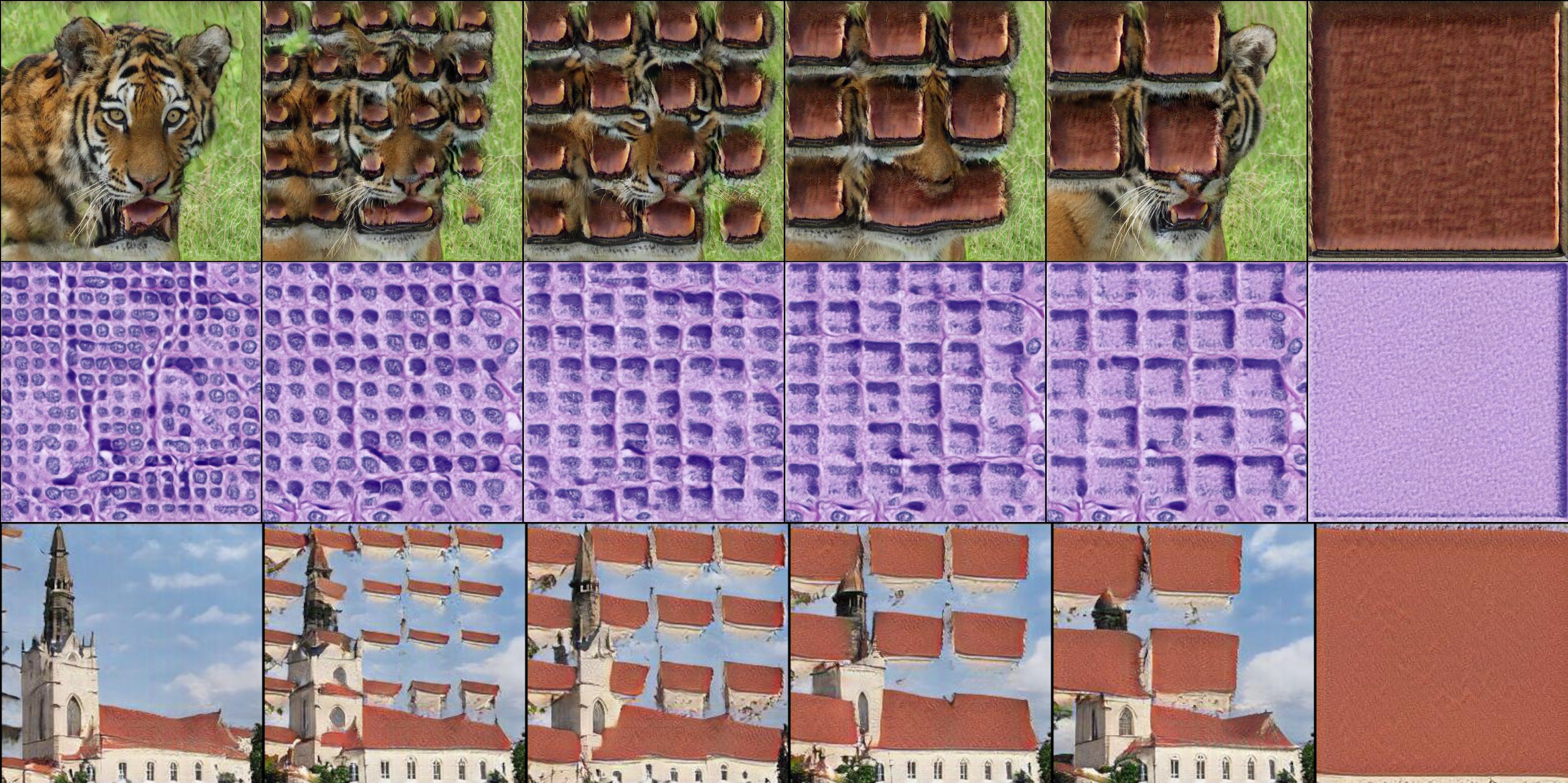}}
\caption{Visualizing non-tileable features with different grid sizes}
\label{fig:vary_grid_size}
\end{center}
\vskip -0.2in
\end{figure*}

\subsubsection{Generation}

We generate 2048 images, and in each image extract a random pixel from LayerX (we choose a layer that is named b16.conv1 in StyleGAN2 and has a spatial size of 16x16) of the GAN.
We generate two visualizations per extracted activation vector $v$: Once by setting all spatial activations to $v$ and once by using a grid of size 2.
Then we compare the activations of both visualizations in Layer2 of the ResNet50 and sort all the 2048 images by their similarity.
Finally Figure \ref{fig:highest_cossim} shows the 4 images with the highest similarity, whereas Figure \ref{fig:lowest_cossim} shows the 4 images with the lowest similarity between the two generated visualizations.

\subsubsection{Measuring Similarity}

In the following we describe in more detail how we measure the similarity between the two visualizations.
First both visualizations get passed till Layer2 of the ResNet50 and we extract their activations at that layer.
Then we multiply those activations by the grid mask used in the GAN to generate the visualization (see Figure \ref{fig:grid_size_visualization}).
Afterwards we compute the mean over height and width, of the activations, weighted by the grid mask.
This gives us two vectors, one per visualization.
Then we compute the cosine similarity between those two vectors.
Finally we show the 4 visualizations with highest cosine similarity in Figure \ref{fig:highest_cossim} and the 4 visualizations with lowest cosine similarity in Figure \ref{fig:lowest_cossim}.

\subsubsection{Effect of Grid Size}

In the following we illustrate the effect of grid size on the visualization.
Figure \ref{fig:vary_grid_size} shows visualizations for non-tileable features using different grid sizes.
The top row shows an example from the GAN trained on AFHQ Wild, the middle row from the GAN trained on BreCaHAD and the bottom row from the GAN trained on LSUN Church.
The grid size for the images from left to right is: 1, 2, 3, 4, 5, full image.
If the grid size is too large, the generated structures stretch too much and become unrealistic.
On the other hand, if the grid size is too small, it may not affect the image enough to show up.
A grid size of 2 or 3 seems to work best.

\begin{figure*}[htb]
\begin{center}
\centerline{\includegraphics[width=\textwidth]{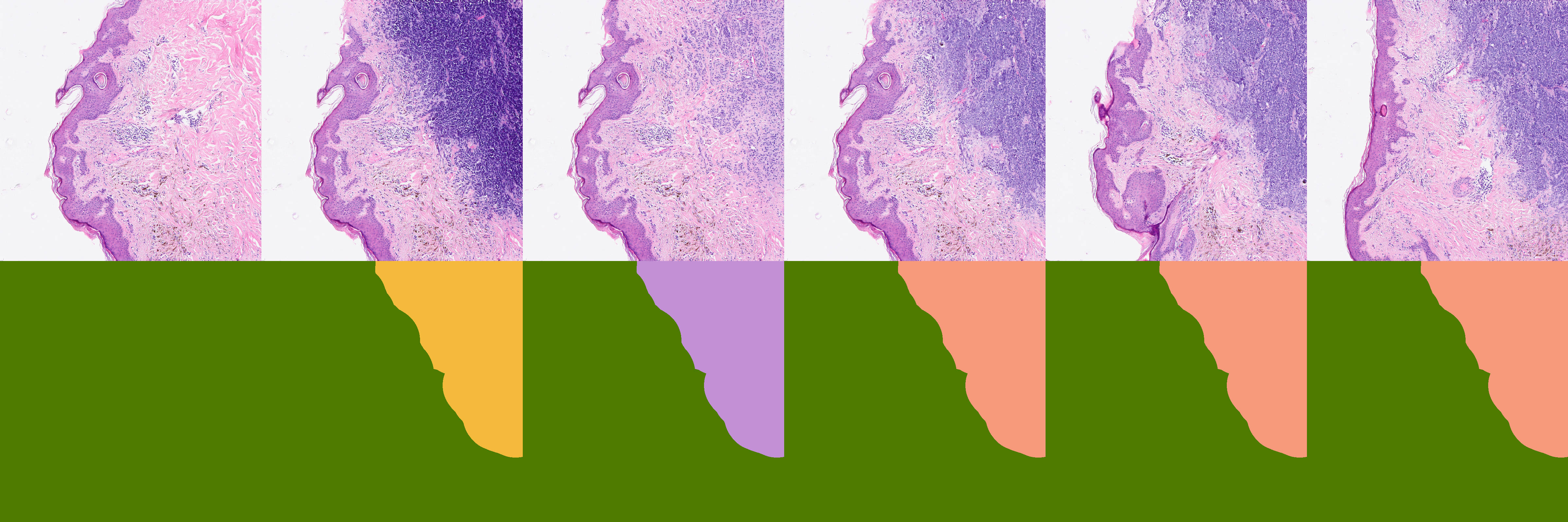}}
\caption{Example for painting with activation vectors, for the digipath GAN. The second from the left and last three images show merkel cell carcinoma, a particular malignant type of cancer}
\label{fig:painting_digipath}
\end{center}
\vskip -0.2in
\end{figure*}

\subsection{Painting with Activation Vectors}

In this section we illustrate a case study for painting with activation vectors.
Figure \ref{fig:painting_digipath} shows the result from painting with activation vectors in a hidden layer of the digipath GAN.
The top row shows the generated images, and the bottom row shows an RGB mask.
The RGB mask is not directly used as input into the model, but each color is mapped to an activation vector from a library and the mask specifies where in the hidden layer the activations should be replaced with the chosen activation vector.
For this the RGB mask gets downsized via nearest neighbour interpolation to the spatial size of the hidden layer (in this case 6x6) and is then used as mask.
The green color in the RGB masks means the activations are not changed there.

In the first four images from the left we keep the noise fixed and only paint with an activation vector.
In the last three images we keep the activation vector fixed and change the noise, resulting in image variations that are still faithful to the RGB mask.
Practically, this method could be used to generate annotated data for semantic segmentation, without needing training data annotated for semantic segmentation.

\subsection{Limitations}

Painting with activation vectors can be problematic, if the GAN has inputs after the painting layer (like the style vector in StyleGAN2).
Because then the structure the activation vector generates also depends on those additional inputs.

In the digipath GAN, we do not have such additional inputs in the decoder part.
But in the training data we have images from different stains, and the activation vectors seem to encode that stain information.
In the normal generation process, we observe stain mixing only in early training of the GAN, but not when it is fully trained (then the whole image is generated in one stain), but we can paint activation vectors from different stains into one image, resulting in more than one stain being present in the image.
But this behavior could also be used as an advantage, by generating combinations of tissues not present in the training data to generate additional training data to increase robustness of a feature extractor, or test its behavior on unusual combinations of diagnoses.

And lastly, painting of non-tileable activation vectors can be difficult to automate (using this painting method to generate annotated data for semantic segmentation, could only work well for tileable activation vectors).

\section{Conclusion}

In this paper we investigated how to visualize activation vectors of hidden layers of GANs, in order to gain insight into how those models represent generated structures internally in their hidden layers.
We observed that the activation vectors correspond to interpretable structures, ranging from e.g. eyes, whiskers and different kinds of fur in the AFHQ GAN, over different kinds of cell nuclei for the BreCaHAD GAN to whole tissue types in the digipath GAN.
Those activation vectors could be divided into two categories: tileable and non-tileable.
Painting with tileable ones tiles the generated structure to fill the painted area, whereas painting with non-tileable activation vectors stretches the generated structure to fill the area, looking unrealistic for larger areas.
%
%
%
Therefore, painting with tileable activation vectors works well, whereas painting larger areas with non-tileable ones is problematic.

For the visualization we recommend generating it twice, once by setting all spatial activations to the activation vector (which works good for tileable structures, but fails for non-tileable ones) and once by setting them in a 2x2 or 3x3 grid (which works good for non-tileable structures, but is a bit harder to interpret for tileable ones).
Finally we note that while we only investigated our methods with GANs, they should also work for other generative models like VAE.

\section{Acknowledgements}
R. Herdt acknowledges the support by the Deutsche Forschungsgemeinschaft (DFG, German Research Foundation) - Projectnumber 459360854 (DFG FOR 5347 Lifespan AI) and the support by the Federal Ministry of Education and Research (BMBF) within the T!Raum-Initiative "\#MOIN! Modellregion Industriemathematik" in the sub-project "MUKIDerm".

\FloatBarrier

\bibliographystyle{IEEEbib}
\bibliography{strings,refs}

\clearpage

\section{Appendix; Investigating Grid Size}

\begin{figure}[htb]
    \centering
    \begin{subfigure}{0.23\textwidth}
        \centering
        \includegraphics[width=\textwidth]{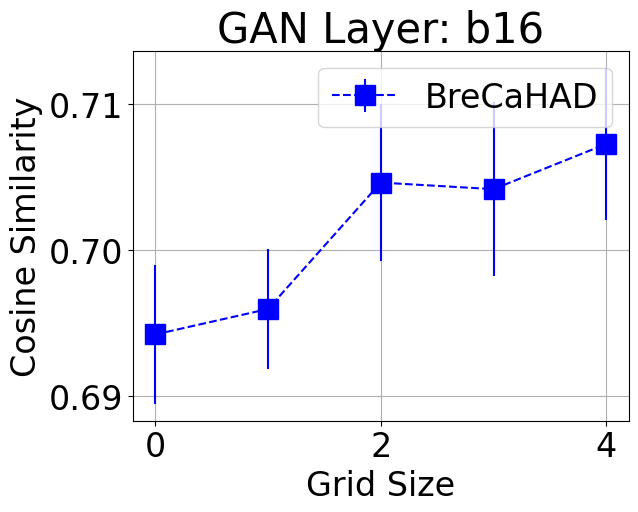}
        \label{fig:sub1}
    \end{subfigure}
    \begin{subfigure}{0.23\textwidth}
        \centering
        \includegraphics[width=\textwidth]{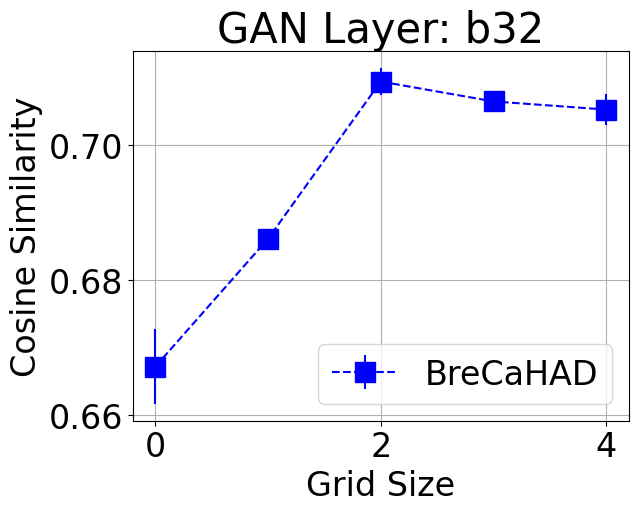}
        \label{fig:sub2}
    \end{subfigure}
    
    \medskip
    
    \begin{subfigure}{0.23\textwidth}
        \centering
        \includegraphics[width=\textwidth]{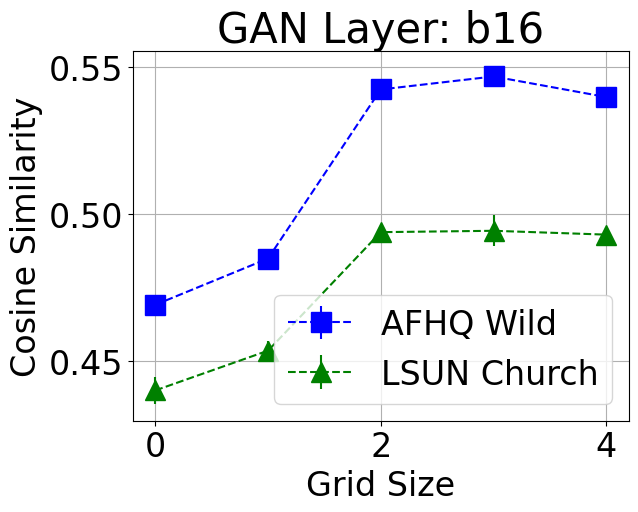}
        \label{fig:sub3}
    \end{subfigure}
    \begin{subfigure}{0.23\textwidth}
        \centering
        \includegraphics[width=\textwidth]{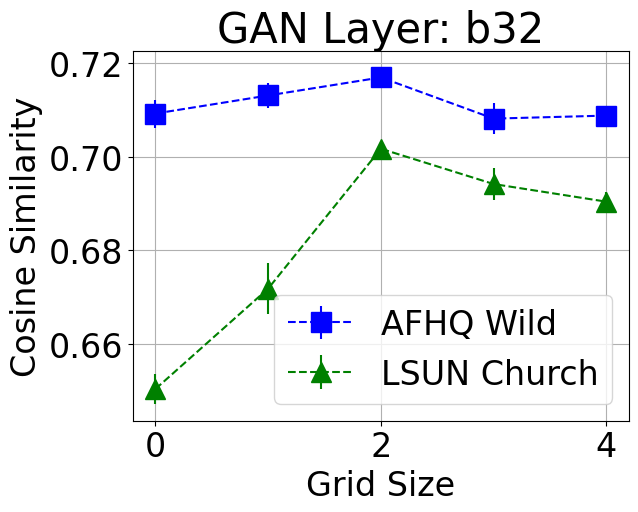}
        \label{fig:sub4}
    \end{subfigure}

    \caption{Cosine similarity of the reconstructed activation vector to the original one. Higher is better.}
    \label{fig:metrics_classifier_pixel}
\end{figure}

In this section we describe an additional experiment regarding which grid size is ideal for the visualization (TLDR a grid size of 2 gives best results).


\begin{algorithm}[htb]
	\caption{Gradient Descent inversion from LayerY}
	\label{alg:invert}
	
		\begin{algorithmic}
            \STATE GAN from layer X to output $G_X$, feature extractor from layers 0 to Y $F_Y$, vectors in LayerX $v^{(1)}$, $v^{(2)}$, vector to be visualized in LayerY $y$

			\FOR{$i \gets 0$ to $n$}
                \STATE $a \gets$ empty tensor of shape $G_X$
    			\STATE blue pixels in $a \gets v^{(1)}_i$
                \STATE white pixels in $a \gets v^{(2)}_i$
                
                \STATE $\text{loss} \gets \nabla_{v^{(1)}_i, v^{(2)}_i} \vert \text{mean}_{h, w}(F_Y(G_X(a))) - y \vert$
    			\STATE backpropagate loss
    			\STATE update $v^{(1)}_i, v^{(2)}_i$
                \ENDFOR	 	
		\end{algorithmic}
\end{algorithm}


We investigate the effect of the grid size (see Figure \ref{fig:grid_size_visualization}), and which grid size works best.
%
%
We generate visualizations for a feature extractor, by inverting it from an activation vector $w$ via gradient descent.
In the following we denote the hidden layer of the feature extractor from which we extract the activation vector $w$ as LayerY.
We do not optimize for the input image, but do the optimization in a hidden layer of a GAN (we use the GAN as preconditioning).
The crucial point is that we do not optimize for all pixels in the hidden layer of the GAN, but only for one or two vectors.
All blue pixels in Figure \ref{fig:grid_size_visualization} would be set to one vector, all white ones to the other and we only optimize for those two vectors (instead of optimizing all spatial activations).
Then we try different grid sizes and measure how good the resulting visualization matches the original activation vector $w$ of the feature extractor.
We run this experiment on the BreCaHAD, AFHQ Wild and LSUN Church GAN.
First we generate 512 images with each of the three GANs.
Then we pass each image till LayerY of the feature extractor and extract a random pixel $y$ at that layer.
After that we invert the feature extractor from $y$, as shown in Algorithm \ref{alg:invert}.
Finally we compute the cosine similarity between the reconstruction $\text{mean}_{h, w}(F_Y(G_X(a)))$ and $y$ and plot the results in Figure \ref{fig:metrics_classifier_pixel} ($\text{mean}_{h, w}$ means computing the mean over height and width).
For each GAN we run the experiment 3 times and then plot the mean and standard deviation.
As feature extractors we use a ResNet50 trained on ImageNet from pytorch model zoo for the AFHQ Wild and LSUN Church GAN, and a ResNet18 trained on Camelyon16 from monai model zoo for the BreCaHAD data.

Figure \ref{fig:metrics_classifier_pixel} shows the cosine similarity between the original and reconstructed activation vector for different grid sizes and GANs.
A grid size of 0 is always worst, which makes sense since there we only optimize for a single vector instead of two (for all other grid sizes it could learn both vectors to be the same, which would be the same as with a grid size of 0).
Overall the cosine similarity is highest with a grid size of 2.

\end{document}